\documentclass{article}
\usepackage[accepted]{tmlr}
\usepackage{url}
\usepackage{booktabs}
\usepackage{graphicx}
\usepackage{amsmath}
\usepackage{amssymb}
\usepackage{placeins}
\usepackage{float}
\usepackage{capt-of}
\usepackage{needspace}
\usepackage[hidelinks]{hyperref}
\hypersetup{
  pdftitle={Auditing GeoFM Evaluation for Field-Extent Segmentation: Label Proxies, Baselines, and When Frozen Features Match Fine-tuning},
  pdfauthor={Santosh Pant},
  pdfsubject={Transactions on Machine Learning Research},
  pdfkeywords={geospatial foundation models, field-extent segmentation, evaluation, fine-tuning, remote sensing}
}
\newcommand{\tablebodygap}{\vspace{0.45em}}

\title{Auditing GeoFM Evaluation for Field-Extent Segmentation: Label Proxies, Baselines, and When Frozen Features Match Fine-tuning}
\author{
\name Santosh Pant\thanks{OpenAI Codex and Anthropic Claude assisted with language editing, software debugging, and consistency checks; the author independently verified the work, made all scientific decisions, and takes full responsibility for it.}
\email pantsantosh23@gmail.com \\
\addr Knox College, Galesburg, Illinois, USA
}
\def\month{08}
\def\year{2026}
\def\openreview{\url{https://openreview.net/forum?id=qRXVTe1yYp}}

\begin{document}
\maketitle

\begin{abstract}
Geospatial foundation-model (GeoFM) benchmarks for agricultural field-extent
segmentation often rely on cropland land-cover proxies and compare against per-pixel
spectral baselines. We measure how these choices affect conclusions using identical
single-date Sentinel-2 pixels from six countries. Replacing polygon-derived field labels
with ESA WorldCover cropland raises random-forest AUROC from 0.55--0.82 to 0.79--0.96,
showing that the proxy creates a more spectrally separable task. On the actual
field-extent labels, a from-scratch U-Net reaches 0.89--0.98 AUROC and has higher point
AUROC than a Prithvi frozen linear probe in five of six regions. We then compare a frozen
backbone and full fine-tuning using the same decoder for Prithvi and TerraMind, with ten matched seeds. At
$\varepsilon{=}0.02$ AUROC, 9 of 12 region$\times$model cells are TOST-equivalent.
Kenya is the common failure case: positives cover roughly 0.1\% of pixels, and full
fine-tuning performs better for both models. Cross-region results are less favorable to
frozen models. On AUROC and average precision (AP), the U-Net leads the full 30-transfer matrix, while an exploratory analysis
excluding Kenya finds that both frozen decoders outperform their fine-tuned counterparts;
no AUROC difference is detected between frozen TerraMind and the U-Net in that subset. For
Prithvi, a frozen decoder trains about $110\times$ fewer parameters than full fine-tuning. These results
argue for auditing labels and baselines before interpreting GeoFM rankings, and for
testing a frozen decoder before assuming that full fine-tuning will help. Our conclusions
are limited to single-date, per-pixel field-extent segmentation, not parcel-boundary or
instance delineation.
\end{abstract}

\section{Introduction}
Evaluating GeoFMs for agricultural field mapping requires two decisions that can change
model rankings: what label defines a field, and what baseline represents supervised
learning. Published studies sometimes use ESA WorldCover ``cropland'' because field
polygons are scarce. Cropland and field extent are related but different targets: one is
a land-cover class, while the other records whether a pixel lies inside a mapped field
polygon. Comparisons also often stop at per-pixel spectral classifiers, even though a
segmentation model can use spatial context.

We audit both decisions on identical Sentinel-2 pixels from six countries in Fields of
The World. To isolate label choice, we score the same models and pixels against polygon-derived
field extent and against WorldCover cropland. To test baseline strength, we compare
per-pixel random forests and frozen GeoFM probes with a U-Net trained from scratch. We then
turn to the adaptation question, using the same decoder, loss, schedule, and matched seeds
for frozen and fully fine-tuned Prithvi and TerraMind backbones.

Label choice changes the benchmark before adaptation enters the comparison. WorldCover
inflates the spectral-baseline score without changing the imagery, whereas polygon-derived
field labels expose the need for a spatial baseline. On those labels, the U-Net has higher
point AUROC than the Prithvi frozen probe in most regions, with India as the exception. PANGAEA
\citep{pangaea} reports a similar supervised-baseline pattern on AI4SmallFarms. The
contrasting comparison in the Prithvi-EO-2.0 report uses multi-temporal crop segmentation
\citep{prithvi} and therefore addresses a different task.

The adaptation comparison also changes once decoder capacity is matched. Frozen and fully
fine-tuned backbones usually agree in-region; Kenya's sparse-positive setting is the shared
failure case. Cross-region transfer is metric-dependent. On AUROC and AP, the U-Net leads
the full 30-transfer matrix. On AUROC, frozen-over-fine-tuned gains appear only in the
exploratory non-Kenya subset. The practical
appeal of a frozen decoder therefore comes from low training cost and frequent in-region
parity. The transfer results do not support a universal advantage.

\paragraph{Contributions.} This paper provides a same-pixel audit of label and baseline
choices in GeoFM field-extent evaluation, a controlled ten-seed comparison of frozen and
fully fine-tuned backbones, and a reproducible protocol that reports both the successful
regions and the sparse-positive Kenya failure case.

\FloatBarrier
\section{Related Work}
\label{sec:related}
\paragraph{Geospatial foundation models.} Recent GeoFMs pretrain on large unlabeled
satellite archives and are reused across tasks: masked-autoencoder models (SatMAE
\citep{satmae}, Scale-MAE \citep{scalemae}, Prithvi-EO \citep{prithvi}), contrastive
and multimodal models (CROMA \citep{croma}, Clay \citep{clay}, TerraMind
\citep{terramind}, AnySat \citep{anysat}), wavelength-conditioned models (DOFA
\citep{dofa}), spectral models (SpectralGPT \citep{spectralgpt}), pixel-timeseries
models (Presto \citep{presto}), and large supervised pretraining (SatlasPretrain
\citep{satlas}). Their downstream evaluations almost always fine-tune the backbone.
We test that default for field-extent segmentation.

\paragraph{Frozen features vs.\ fine-tuning.} Outside Earth observation, a
substantial literature documents that frozen pretrained features with a light head
are often competitive with full fine-tuning, and that fine-tuning can harm
out-of-distribution performance. \citet{kumar2022finetuning} show that fine-tuning
distorts pretrained features and underperforms frozen linear probing under
distribution shift, which is the pattern we observe across regions. Locked-feature
training \citep{zhai2022lit}, transfer studies
\citep{kornblith2019transfer,kolesnikov2020bit}, and the linear-probe evaluation of
self-supervised models \citep{he2022mae} all point in the same direction. We
find evidence consistent with the same pattern for geospatial field-extent
segmentation, with an explicit boundary condition and a cost analysis.

\paragraph{GeoFM-specific evaluation questions.}
Prior work establishes that frozen representations can remain competitive under
distribution shift. It does not resolve how that finding translates to GeoFM benchmarks,
where a frozen backbone is often given only a linear head while its fine-tuned counterpart
uses a decoder. We remove this capacity difference by giving each frozen and fine-tuned
pair the same decoder (2.75M parameters for Prithvi; 2.16M for TerraMind). This
capacity-matched comparison tests whether a reported fine-tuning gain comes from
backbone adaptation or from giving the fine-tuned model a stronger decoder. With
decoder capacity matched, the difference falls within the
declared equivalence margin in 9 of 12 cells (Section~\ref{sec:spine}).

The available labels introduce a second geospatial question that is not captured by
the in-distribution/out-of-distribution framing in \citet{kumar2022finetuning}: what
task does the benchmark measure, and what happens when positives are exceptionally
sparse? On the same pixels,
WorldCover cropland raises the per-pixel random forest from 0.55--0.82 to 0.79--0.96
AUROC (Table~\ref{tab:flaws}), so a land-cover proxy changes the task being measured.
Kenya marks the limit of the frozen-decoder result: at roughly 0.1\% positive pixels and
155 training chips, the frozen decoder trails full fine-tuning by 0.10--0.12 AUROC for
both GeoFMs and produces near-all-negative predictions at threshold 0.5
(Section~\ref{sec:boundary}). These conditions make the result a geospatial evaluation
finding about where a frozen decoder works and fails. They do not establish a new
general claim about frozen representations.

\paragraph{EO benchmarks and evaluation.} GEO-Bench \citep{geobench} and PANGAEA
\citep{pangaea} standardize GeoFM evaluation; we use PANGAEA's results as external
corroboration. Fields of The World \citep{ftw} supplies the field polygons
that let us replace land-cover proxies. Spatial autocorrelation is a well-known
evaluation pitfall in geospatial ML: train and test
chips drawn from nearby tiles can inflate scores. We use a chip-grouped split
(no chip in both partitions) and report tile-disjoint sensitivity
(Appendix Table~\ref{tab:tilesplit}) to address it. Our subject is the evaluation
design, specifically labels and baselines; we do not propose a new model.

\FloatBarrier
\section{Setup}
\label{sec:setup}
\paragraph{Data.} We use Sentinel-2 L2A chips paired with field polygons from
Fields of The World \citep{ftw} across six countries spanning smallholder to
industrial agriculture: India, Cambodia, Vietnam, Kenya, France, and the Netherlands.
For each chip we form two labels on the same pixels. The \textsc{true} label is 1
when the pixel lies inside an FTW field polygon, rasterized to the chip grid in the
chip's own CRS. The \textsc{proxy} label is 1 when the pixel is ESA WorldCover
``cropland'' (class~40, \citealp{worldcover}). The task throughout is per-pixel binary
field-extent segmentation (inside-vs-outside a field polygon); we do not evaluate
boundary delineation or instance separation between adjacent parcels, and we use
``field mapping'' as a synonym for this per-pixel task.

\paragraph{Models.} Non-FM baselines: a per-pixel random forest on the 12 raw S2
bands, a $5\times5$ windowed random forest (300-d), and a from-scratch U-Net
\citep{unet} (31M parameters). GeoFMs: Prithvi-EO-2.0-300M \citep{prithvi} and
TerraMind \citep{terramind} (with Clay and AnySat for breadth), used in two modes.
Frozen: backbone weights fixed, then either a linear probe or a trained decoder is fit on frozen features.
Fine-tuned: backbone trained end-to-end with the same decoder.

\paragraph{Nomenclature.} We use three terms consistently. \textbf{Frozen-probe} =
backbone weights fixed, a linear classifier (${\sim}10^3$ parameters) trained on
extracted per-pixel features; Table~\ref{tab:flaws} reports Prithvi consistently,
with the four-model breakdown in Appendix Table~\ref{tab:fm_breakdown}.
\textbf{Frozen-decoder} = backbone
weights fixed, a 4-stage convolutional decoder (2.75M parameters for Prithvi;
2.16M for TerraMind) trained on
extracted feature maps; this is the apples-to-apples comparator against full
fine-tuning and is the configuration used throughout
Sections~\ref{sec:spine}--\ref{sec:cost} unless noted. \textbf{Full-FT} = backbone
and the same decoder trained jointly end-to-end. For Prithvi, this configuration
has 307M trainable parameters. The
three configurations can disagree: in Kenya, the best frozen probe across four FMs
reports 0.853 AUROC (AnySat; separate frozen-probe sensitivity), while the ten-seed
frozen-decoder on Prithvi reports 0.653; we report
both with the configuration explicit and discuss the gap in
Section~\ref{sec:boundary}.

\begin{table}[!htb]\centering\footnotesize
\caption{\textbf{Model configurations used in this paper.} Five configurations
appear across the results tables; each row names the trainable component,
trainable-parameter count, whether the backbone is updated, the training budget,
and which tables report each configuration. Decoder counts use Prithvi/TerraMind order.}
\label{tab:configs}
\begin{tabular}{lrcll}
\toprule
Configuration & Trainable params & Backbone updated & Optimizer / epochs & Reported in \\
\midrule
per-pixel / windowed RF & (fitted trees) & n/a & sklearn defaults & Tables~\ref{tab:flaws},~\ref{tab:edgezone} \\
frozen-probe (GeoFM $+$ linear) & ${\sim}10^{3}$ & no & logistic regression & Tables~\ref{tab:flaws},~\ref{tab:edgezone},~\ref{tab:fm_breakdown} \\
frozen-decoder (GeoFM $+$ 4-stage conv) & $2.75$M / $2.16$M & no & AdamW, 80 epochs & Tables~\ref{tab:spine},~\ref{tab:iou_summary},~\ref{tab:paired_tost} \\
full-FT (GeoFM $+$ same decoder) & $307$M (Prithvi) & yes & AdamW, 80 epochs & Tables~\ref{tab:spine},~\ref{tab:iou_summary},~\ref{tab:paired_tost} \\
U-Net (from scratch) & $31$M & n/a & Adam, 150 epochs & Tables~\ref{tab:flaws},~\ref{tab:spine},~\ref{tab:iou_summary} \\
\bottomrule
\end{tabular}
\end{table}
\paragraph{Protocol and statistics.} All models are scored on the same held-out test
pixels via a chip-grouped split (no chip in both train and test; overlap $=0$; seed
20260514, 25\% test). The main frozen-decoder, full-FT, and U-Net comparisons use a
single-recipe ten-seed grid with ten matched training seeds per region and one software recipe;
we report mean and a seed-paired $t$-interval with $\mathrm{df}=9$ for matched
frozen-decoder vs.\ full-FT differences (the same paired analysis used in
Table~\ref{tab:spine} and Appendix Table~\ref{tab:paired_tost}). Frozen-probe AUROC in
Table~\ref{tab:flaws} carries chip-level 95\% bootstrap CIs ($B{=}1000$) recomputed
over the test chips, available in five regions (Cambodia per-chip CIs not
recomputed; point estimates only). Equivalence claims use two one-sided tests
(TOST; \citealp{lakens2017equivalence}) with declared margin
$\varepsilon{=}0.02$ AUROC in-region; Holm correction across the 12 primary TOST
p-values retains all 9 equivalent cells at $\alpha{=}0.05$
(\verb|ftw_inregion_equivalence.json|).

\paragraph{TOST margin justification.} We use $0.02$ AUROC as a practical tolerance,
not as a universal standard for field mapping. It is less than half of the India
frozen-probe-over-U-Net gap ($0.042$) that we interpret substantively in Section~\ref{sec:flaws},
and more than twice the largest per-configuration seed
standard deviation outside the three failing cells ($0.0085$). The sensitivity analysis
in Appendix Table~\ref{tab:eps_sweep} gives the result across margins from $0.005$ to
$0.030$. The failing set is unchanged from $0.020$ through $0.030$. At $0.015$,
Prithvi--Netherlands also fails; at $0.010$, TerraMind--Netherlands joins it. Both point
estimates favor the frozen decoder. At $0.005$, Prithvi--India and TerraMind--Cambodia also fail;
their fine-tuning-favored differences are $0.004$ and $0.006$ AUROC.

\paragraph{Cross-region uncertainty and reproduction.} Cross-region
uncertainty is seed-paired over ten seeds, with each seed contributing its mean
over the directed-transfer subset. The multiple-comparison family contains 10
directional tests: two claim families in each of the five regions with retained
per-chip uncertainty. Cambodia is excluded because its per-chip CIs were not
recomputed. The tests compare RF\textsubscript{proxy}$-$RF\textsubscript{true}
on F1 and best-FM$-$RF\textsubscript{true} on AUROC. All 10 survive both Benjamini--Hochberg FDR
control at 0.05 and Holm correction at 0.05, with minimum $z{=}6.54$
(\verb|ftw_multiple_comparison.json|); chip-paired Wilcoxon signed-rank confirms
the India Prithvi-over-U-Net exception ($p{=}2.4\times10^{-4}$, all 12 chunks
positive). The evaluation code, metric artifacts, and reproduction commands are
released. Primary aggregates recompute from the released run files, and direct
raster checks verify label-to-image alignment (Appendix~\ref{app:repro}).

\FloatBarrier
\section{Two Evaluation Choices that Change the Result}
\label{sec:flaws}
All GeoFM entries in this section are the frozen-probe configuration (Prithvi linear probe on extracted per-pixel features) unless labeled `FT'; see Table~\ref{tab:configs}.
Table~\ref{tab:flaws} reports, per region, the per-pixel RF and the strong models
under both labels (\textsc{true}-label AUROC unless noted).

\begin{table}[H]
\centering\small
\caption{\textbf{Two evaluation choices (true-label AUROC; six regions).} The
WorldCover proxy lifts the per-pixel RF from 0.55--0.82 to 0.79--0.96, making it
foundation-model-competitive. A from-scratch U-Net on \textsc{true} labels reaches
0.89--0.98 everywhere and is competitive with Prithvi frozen-probe in five of six
regions (India excepted), so the reported ``GeoFM\,$\gg$\,baseline'' gap mostly
reflects the choice of baseline. The Prithvi frozen-probe column reports a single
consistent model (linear probe on Prithvi-EO-2.0 features); a per-model breakdown
across \{Prithvi, TerraMind, Clay, AnySat\} is in Appendix Table~\ref{tab:fm_breakdown}.
Bracketed values are 95\% CIs: chip-level bootstrap ($B{=}1000$) for the
frozen-probe column (five regions; Cambodia point estimate only); seed-level
$t$-interval ($\mathrm{df}=9$) for ten-seed U-Net and Prithvi full-FT. The full-FT
column is end-to-end fine-tuned Prithvi. Per-model tile-disjoint sensitivity is in
Appendix Table~\ref{tab:tilesplit}. The `Prithvi frozen-probe' column is a
linear probe on Prithvi-EO-2.0 features (\emph{not} the frozen-decoder
configuration used in Table~\ref{tab:spine}); see Table~\ref{tab:configs} for
the full configuration key.}
\label{tab:flaws}
\tablebodygap
\begin{tabular}{lccccc}
\toprule
Region & RF \textsc{true} & RF \textsc{proxy} & U-Net & Prithvi frozen-probe & Prithvi full-FT \\
\midrule
India        & 0.574 & 0.786 & 0.941\,{\scriptsize[.925,.958]} & 0.983\,{\scriptsize[.979,.985]} & 0.985\,{\scriptsize[.984,.987]} \\
Cambodia     & 0.550 & 0.900 & 0.948\,{\scriptsize[.947,.949]} & 0.924\,{\scriptsize(no CI)}            & 0.949\,{\scriptsize[.949,.949]} \\
Vietnam      & 0.643 & 0.911 & 0.964\,{\scriptsize[.961,.967]} & 0.929\,{\scriptsize[.925,.933]} & 0.958\,{\scriptsize[.958,.959]} \\
Kenya        & 0.651 & 0.806 & 0.891\,{\scriptsize[.885,.897]} & 0.766\,{\scriptsize[.718,.813]} & 0.771\,{\scriptsize[.738,.804]} \\
France       & 0.695 & 0.959 & 0.977\,{\scriptsize[.977,.978]} & 0.961\,{\scriptsize[.958,.964]} & 0.979\,{\scriptsize[.978,.980]} \\
Netherlands  & 0.815 & 0.928 & 0.981\,{\scriptsize[.980,.983]} & 0.955\,{\scriptsize[.951,.958]} & 0.958\,{\scriptsize[.953,.963]} \\
\bottomrule
\end{tabular}
\end{table}
\FloatBarrier

\paragraph{Choice 1: label source.} The proxy raises the
per-pixel RF by $+0.11$ to $+0.35$ AUROC in every region, turning a near-chance
field-membership predictor into one that rivals GeoFMs. The reason is that WorldCover
``cropland'' is spectrally separable while true field membership is a spatial
property. Across the five regions with shipped coverage, WorldCover labels 22--70\% of
pixels as cropland, versus 0.1--18\% true field coverage. The proxy therefore asks ``is this managed land?''
instead of ``is this pixel in a field?''

\paragraph{Choice 2: baseline strength.}
On \textsc{true} labels a supervised U-Net reaches 0.89--0.98 AUROC in all six
regions. Compared against a single consistent FM (Prithvi frozen-probe), the U-Net
has higher point AUROC in five of six regions (Vietnam, Kenya, France, Netherlands,
Cambodia); India is the only GeoFM exception (Wilcoxon $p{=}2.4\times10^{-4}$
on per-chip paired chunks). For Vietnam, the U-Net's ten-seed CI $[.961,.967]$
is disjoint from frozen-Prithvi's $[.925,.933]$; for Cambodia we have only point
estimates (no per-chip CI). India is the only region where frozen-Prithvi's CI
$[.979,.985]$ is above the U-Net point estimate (0.941). The
``GeoFM\,$\gg$\,baseline'' gap therefore depends substantially on the choice
of baseline; by itself, it is not evidence of a transferable representational advantage. A per-pixel RF
cannot use spatial context, and a $5\times5$ windowed RF closes only about 0.01 of
the gap (with a small drop on Kenya; Table~\ref{tab:windowed}). The per-region best across \{Prithvi, TerraMind, Clay, AnySat\}
(Appendix Table~\ref{tab:fm_breakdown}) raises the frozen-FM Kenya and Netherlands
cells to 0.853 (AnySat) and 0.961 (AnySat), but tile-disjoint sensitivity
(Appendix Table~\ref{tab:tilesplit}) shows the AnySat advantage does not survive a
split-design change; on the apples-to-apples Prithvi-only frozen-probe comparison
reported above, the U-Net is the stronger model in those two regions.

\paragraph{India exception.} In-region,
Prithvi frozen-probe beats the U-Net in only one of six regions (India), where
Prithvi leads by 0.042 AUROC, and the direction is supported by a separate
chip-paired one-sided Wilcoxon
signed-rank test ($n{=}12$ paired chunks; $p{=}2.4\times10^{-4}$; all 12 pairs
positive). The chunk audit shows that India is stable across test chunks rather than
a seed artifact, so we retain it as an exception. Possible contributors include Prithvi's HLS global pretraining
(which may include Indian agro-climates in unknown proportion), India's
distinctive smallholder spatial layout, and ordinary single-region noise at the
between-region scale. The released alignment summary gives Cambodia a smaller
mean polygon area than India (0.20 vs.\ 0.36 ha), but the U-Net
leads its frozen probe (0.948 vs.\ 0.924), so field size alone does not explain the pattern. We do not have a
single confident explanation. The India exception cautions against treating the
supervised-baseline result as universal; the field-size comparisons above do not identify a mechanism.

\paragraph{Training-budget control.} The U-Net comparison is not epoch-matched to the GeoFM full-FT runs (150 epochs
from-scratch vs.\ 80 from a pretrained backbone). An epoch-matched U-Net control
at 80 epochs is reported in Appendix Table~\ref{tab:unet_epoch_sensitivity}.
At the matched 80-epoch budget the U-Net still has higher point AUROC than the
Prithvi frozen probe in five of six regions (India remains the same exception by
point estimate), and the
80-epoch means are within $0.01$ AUROC of the 150-epoch means in five of six
regions; India shows the largest drop ($0.941 \to 0.911$, seed std $0.023$),
which widens the existing India frozen-probe lead over the U-Net but does not
create a new exception. Epoch count is also a poor proxy for compute here: our
approximate development-run observations were ${\approx}15$ GPU-min per region
for the 150-epoch U-Net and ${\approx}20$ for 80-epoch full-FT
(Table~\ref{tab:cost}; elapsed time was not retained in the result JSONs). The primary
frozen-vs-fine-tuning comparison in Section~\ref{sec:spine} is separately
budget-matched by construction (both 80 epochs, identical AdamW schedule and
matched seeds; Table~\ref{tab:recipe}).
Together with the 150-epoch convergence diagnostic (Appendix
Figure~\ref{fig:convergence}), these controls do not indicate that the U-Net result
is an artifact of under-training the GeoFMs.

\paragraph{External corroboration (PANGAEA).} On AI4SmallFarms \citep{ai4smallfarms}
(manually digitized field boundaries), PANGAEA \citep{pangaea} reports a from-scratch
U-Net at 46.3 mIoU, beating every evaluated GeoFM (Prithvi 26.9, DOFA 27.1, Scale-MAE
21.5). Conversely, the Prithvi-EO-2.0 result of $+8.1$ mIoU over a from-scratch
U-Net on multi-temporal crop segmentation \citep{prithvi} comes from a crop-class
segmentation task rather than a polygon-derived field-extent benchmark. Both evaluation choices appear in the published literature,
independent of our evaluation setup.

\FloatBarrier
\Needspace{4\baselineskip}
\section{Frozen Decoders Often Match Full Fine-tuning In-Region; Cross-Region Depends on Regime}
\label{sec:spine}
Except for the frozen-probe boundary-zone control in Table~\ref{tab:edgezone}, all GeoFM
comparisons in this section use the frozen-decoder and full-FT configurations with matched training seeds; see Table~\ref{tab:configs}.
Holding the decoder, recipe, labels, and evaluation pixels fixed, Table~\ref{tab:spine}
compares frozen-backbone GeoFM features against full end-to-end fine-tuning in a
single-recipe ten-seed grid with ten matched seeds.

\begin{table}[!htb]
\centering\footnotesize
\caption{\textbf{Frozen-decoder vs.\ full fine-tuning, in-region and
cross-region (true-label AUROC).}
\emph{Left:} frozen backbone + trained decoder vs.\ full fine-tuning of the same
architecture, with ten matched seeds per cell. 95\% CIs on
$\Delta=\mathrm{frozen}-\mathrm{full\mbox{-}FT}$ use a seed-paired $t$-interval
($\mathrm{df}=9$). A dagger marks TOST-equivalence at $\varepsilon{=}0.02$ AUROC by the
conservative criterion that the 95\% CI lies fully within $[-0.02,+0.02]$:
9 of 12 cells pass. Kenya rejects equivalence for both GeoFMs with full-FT ahead;
TerraMind India is non-equivalent in the frozen-favored direction because the CI
extends past $+0.02$. \emph{Right:} mean AUROC over all 30 directed transfers and
the exploratory non-Kenya 20 subset (see Section~\ref{sec:spine} for provenance).
  U-Net has the highest AUROC on the full matrix; on exploratory non-Kenya transfers,
  both frozen GeoFMs beat their full-FT counterparts, and the TerraMind-frozen-minus-U-Net
  AUROC CI includes zero.}
\label{tab:spine}
\tablebodygap
\begin{minipage}[t]{0.59\linewidth}
\vspace{0pt}\centering\scriptsize
\textbf{In-region: frozen decoder vs.\ full-FT}\\[3pt]
\setlength{\tabcolsep}{2pt}
\begin{tabular*}{\linewidth}{@{\extracolsep{\fill}}llcccc@{}}
\toprule
Model & Region & frozen & full-FT & $\Delta$ & 95\% CI \\
\midrule
Prithvi & India & 0.982 & 0.985 & $-0.004$ & $[-0.006,-0.002]^\dagger$ \\
Prithvi & Cambodia & 0.947 & 0.949 & $-0.002$ & $[-0.003,-0.001]^\dagger$ \\
Prithvi & Vietnam & 0.955 & 0.958 & $-0.004$ & $[-0.004,-0.003]^\dagger$ \\
Prithvi & Kenya & 0.653 & 0.771 & $-0.117$ & $[-0.176,-0.059]$ \\
Prithvi & France & 0.982 & 0.979 & $+0.003$ & $[+0.002,+0.004]^\dagger$ \\
Prithvi & Netherlands & 0.971 & 0.958 & $+0.013$ & $[+0.008,+0.018]^\dagger$ \\
\midrule
TerraMind & India & 0.968 & 0.948 & $+0.020$ & $[+0.005,+0.036]$ \\
TerraMind & Cambodia & 0.942 & 0.948 & $-0.006$ & $[-0.007,-0.005]^\dagger$ \\
TerraMind & Vietnam & 0.951 & 0.956 & $-0.004$ & $[-0.005,-0.004]^\dagger$ \\
TerraMind & Kenya & 0.613 & 0.710 & $-0.097$ & $[-0.146,-0.048]$ \\
TerraMind & France & 0.979 & 0.978 & $+0.001$ & $[+0.000,+0.002]^\dagger$ \\
TerraMind & Netherlands & 0.961 & 0.952 & $+0.009$ & $[+0.003,+0.015]^\dagger$ \\
\bottomrule
\end{tabular*}
\end{minipage}\hfill
\begin{minipage}[t]{0.38\linewidth}
\vspace{0pt}\centering\scriptsize
\textbf{Cross-region mean AUROC}\\[3pt]
\setlength{\tabcolsep}{2.5pt}
\begin{tabular*}{\linewidth}{@{\extracolsep{\fill}}lcc@{}}
\toprule
Config & all 30 & expl.\ non-Kenya 20 \\
\midrule
U-Net (scratch)    & \textbf{0.826} & \textbf{0.842} \\
TerraMind frozen-dec. & 0.781 & 0.839 \\
TerraMind full-FT  & 0.781 & 0.813 \\
Prithvi full-FT    & 0.780 & 0.810 \\
Prithvi frozen-dec. & 0.767 & 0.816 \\
\bottomrule
\end{tabular*}
\end{minipage}
\par\vspace{0.35em}
{\footnotesize $^\dagger$~TOST-equivalent within $\pm 0.02$ AUROC
(seed-paired $t$-interval, $\mathrm{df}=9$, $n{=}10$ seeds).}
\end{table}

\begin{table}[H]\centering\small
\caption{\textbf{In-region IoU complements AUROC} (mean over 10 seeds,
threshold $p{=}0.5$). The IoU pattern favors the U-Net more often than AUROC:
U-Net wins outright in five regions (India, Vietnam, Kenya, France, Netherlands),
while Full-FT leads in Cambodia. The Kenya entries for the frozen and
full-FT configurations remain near all-negative at threshold 0.5. Full
AUROC/AP/F1/IoU breakdown in Appendix Table~\ref{tab:xmetrics}.}
\label{tab:iou_summary}
\tablebodygap
\begin{tabular}{lccc}
\toprule
Region (positive \%) & U-Net IoU & Frozen-decoder IoU & Full-FT IoU \\
\midrule
India ($\sim$0.3\%)        & \textbf{0.455} & 0.386 & 0.433 \\
Cambodia ($\sim$23\%)      & 0.826 & 0.850 & \textbf{0.856} \\
Vietnam ($\sim$18\%)       & \textbf{0.854} & 0.839 & 0.847 \\
Kenya ($\sim$0.1\%)        & \textbf{0.086} & 0.000 & 0.005 \\
France ($\sim$11\%)        & \textbf{0.891} & 0.850 & 0.856 \\
Netherlands ($\sim$3.8\%)  & \textbf{0.809} & 0.699 & 0.708 \\
\bottomrule
\end{tabular}
\end{table}
\FloatBarrier

\paragraph{In-region.} With the backbone frozen and only a decoder trained,
performance is statistically equivalent to full fine-tuning in 9 of 12
region$\times$model cells across the single-recipe ten-seed grid (Table~\ref{tab:spine},
TOST at $\varepsilon{=}0.02$ AUROC; count stable for $\varepsilon \in [0.02, 0.03]$;
no FT-favored non-equivalence besides Kenya at any $\varepsilon \geq 0.010$;
full per-seed breakdown in Appendix Table~\ref{tab:paired_tost} and sensitivity in
Appendix Table~\ref{tab:eps_sweep}). Prithvi is equivalent outside Kenya:
India, Cambodia, Vietnam, France, and the Netherlands. TerraMind is equivalent in
four of six: Cambodia, Vietnam, France, and the Netherlands. The three
non-equivalent cells have interpretable directions: Kenya favors full-FT for
both GeoFMs (Prithvi $\Delta{=}{-}0.117$, CI $[-0.176,-0.059]$; TerraMind
$\Delta{=}{-}0.097$, CI $[-0.146,-0.048]$), while TerraMind India favors the
frozen decoder in direction but misses equivalence because the CI extends past the
$+0.02$ margin ($\Delta{=}{+}0.020$, CI $[+0.005,+0.036]$). Outside Kenya, the
largest fine-tuning advantage is TerraMind Cambodia at $0.006$ AUROC, well inside
the declared equivalence band. An earlier apparent fine-tuning advantage came from
an architecture confound: comparing a fine-tune-plus-decoder against a frozen
linear probe. Giving the frozen backbone the same decoder removes the gap in most
in-region cells.

\paragraph{IoU complements AUROC.} For an imbalanced segmentation task (positive
rates spanning 0.1\%--23\%), AUROC alone can obscure threshold behavior.
Table~\ref{tab:iou_summary} reports mean per-region IoU at threshold 0.5. The U-Net
leads in five regions, while full-FT leads only in Cambodia. The Kenya entries for the
frozen and full-FT configurations remain near all-negative at this threshold. Thus the
segmentation metric favors the U-Net more often than AUROC does, but it does not show a
systematic full-FT advantage over the frozen decoder. We do not make an IoU-equivalence
claim.

\paragraph{Boundary-zone scope.} Our task is field-extent
segmentation, not boundary delineation or instance separation
(Section~\ref{sec:setup}). As a control on the frozen-probe comparisons in
Table~\ref{tab:flaws}, we re-evaluated the frozen-probe feature classifiers using a
boundary-zone pixel pool within $\le 2$ pixels of a field edge over the India chips
($n{=}203{,}864$), split 75/25 by chip. AUROC is computed only on the held-out split. The five classifiers evaluated here (four frozen probes $+$ the per-pixel
RF) all fall to near-chance on this slice (Table~\ref{tab:edgezone}: AUROC
0.57--0.59 for Clay, Prithvi, TerraMind, AnySat, RF), against 0.94--0.98 on the
full-region task for the three transformer-based FMs. We did not evaluate the U-Net, frozen-decoder, or full-FT
configurations on this slice, so Table~\ref{tab:edgezone} supports \emph{only} a
frozen-probe scope claim: it does not rank the trained segmentation configurations
on boundary pixels. Papers using FTW for true boundary delineation (e.g.\ object-F1
or Boundary-IoU style metrics) ask a strictly harder question and are outside our
scope.

\begin{table}[H]\centering\small
\caption{\textbf{Frozen-probe boundary-zone evaluation (India).} A pool of
$n{=}203{,}864$ pixels within $\le 2$ pixels of a field edge is split 75/25 by chip,
and AUROC is computed only on the held-out split. The five classifiers evaluated here
(four frozen probes $+$ the per-pixel RF) all fall to near-chance AUROC, against
0.94--0.98 on the full-region task for the three transformer-based FMs.
\textbf{Scope.} The U-Net, frozen-decoder, and full-FT configurations were
\emph{not} evaluated on this slice; this table supports only a frozen-probe
scope claim and does not compare trained segmentation models on boundary
pixels. None of the four frozen-probe configurations is a boundary-delineation
specialist; the frozen-probe field-extent advantage rests on interior context.}
\label{tab:edgezone}
\tablebodygap
\begin{tabular}{lcccccc}
\toprule
Slice & RF & Clay & Prithvi & TerraMind & AnySat & full-region AUROC reference \\
\midrule
boundary zone ($\le$2 px) & 0.586 & 0.571 & 0.576 & 0.584 & 0.582 & see row below \\
full region (India)       & 0.574 & 0.577 & 0.983 & 0.967 & 0.939 & cf.\ Table~\ref{tab:flaws} \\
\bottomrule
\end{tabular}
\end{table}
\FloatBarrier

\paragraph{Cross-region.} For cross-country transfer without target-region
labels, the ten-seed grid covers all 30 directed train-country$\to$test-country
transfers among the six regions (Appendix Table~\ref{tab:xfer-full}). The full
matrix does \emph{not} support a broad frozen-GeoFM transfer win: the from-scratch
U-Net has the highest mean AUROC (0.826), followed by TerraMind full-FT and
TerraMind frozen (both 0.781), Prithvi full-FT (0.780), and Prithvi frozen
(0.767). Seed-paired deltas over the 30 transfers show U-Net ahead of both frozen
GeoFMs (Prithvi frozen minus U-Net $=-0.059$, CI $[-0.067,-0.051]$; TerraMind
frozen minus U-Net $=-0.045$, CI $[-0.053,-0.036]$). Prithvi frozen also trails
Prithvi full-FT on the full matrix ($-0.013$, CI $[-0.021,-0.006]$), while
the TerraMind comparison detects no difference ($-0.000$, CI
$[-0.013,+0.013]$).

The non-Kenya subset was not pre-registered; we defined it after observing Kenya's
in-region behavior and treat it as exploratory. The full 30-transfer matrix remains
the confirmatory cross-region analysis.
Kenya is
independently characterized as a sparse-positive case on three grounds before its
transfers are partitioned out: extreme label sparsity ($\sim$0.1\% positive
pixels, the lowest in the study), 155 training chips, and
threshold-metric collapse on trained GeoFM configurations (Section~\ref{sec:boundary},
Appendix Table~\ref{tab:xmetrics}). On the exploratory non-Kenya 20 directed transfers, both
frozen GeoFMs beat their full-FT counterparts: Prithvi frozen minus Prithvi full-FT
is $+0.006$ AUROC (CI $[+0.001,+0.011]$), and TerraMind frozen minus TerraMind
full-FT is $+0.025$ (CI $[+0.019,+0.032]$). Relative to the U-Net, the TerraMind
frozen CI includes zero on exploratory non-Kenya transfers
($-0.004$, CI $[-0.013,+0.005]$), while Prithvi frozen is lower
($-0.026$, CI $[-0.033,-0.019]$). On the full matrix, a from-scratch U-Net has the
highest AUROC and AP.
The IoU@0.5 ranking differs: TerraMind frozen exceeds the U-Net by 0.02 on the full
matrix and by 0.05 on the exploratory subset (Appendix Table~\ref{tab:xfer_metrics}).
The U-Net comparison is therefore specific to AUROC and AP. Under the thresholded
segmentation metric, frozen TerraMind is competitive with or better than the U-Net.
A task-trained supervised model also remains far stronger than a per-pixel spectral
baseline (Table~\ref{tab:flaws}).

\begin{figure}[H]
\centering
\includegraphics[width=0.6\linewidth]{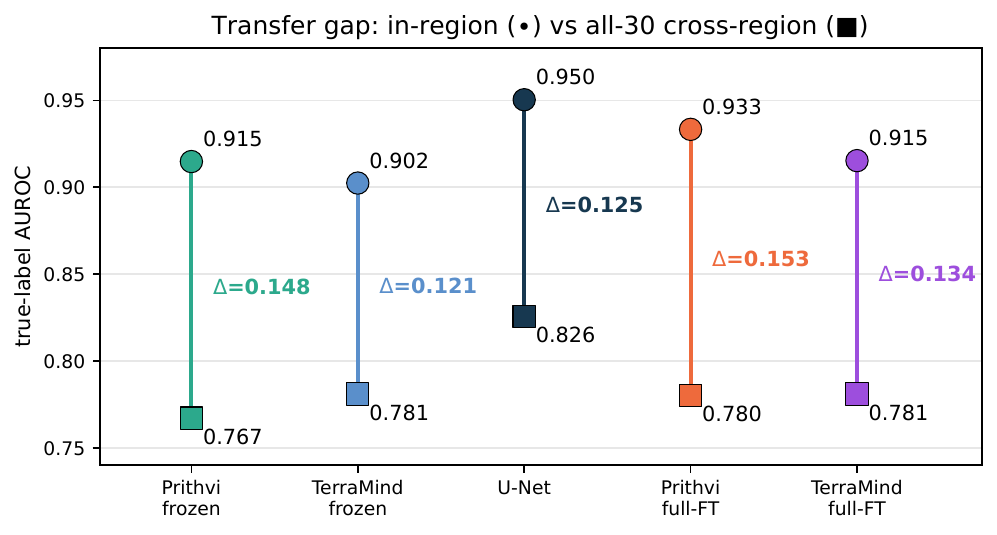}
\caption{\textbf{Transfer gap.} Six-region in-region mean ($\bullet$) vs.\
full 30-transfer cross-region mean ($\blacksquare$). The U-Net has the highest AUROC
on the full matrix; frozen decoders serve mainly as lower-cost alternatives to full
fine-tuning, not as general cross-region winners.}
\label{fig:gap}
\end{figure}

\FloatBarrier
\Needspace{4\baselineskip}
\section{Characterizing the Kenya Sparse-Positive Regime}
\label{sec:boundary}
The results in this section use the same frozen-decoder and full-FT configurations as Section~\ref{sec:spine}; see Table~\ref{tab:configs}.
Kenya is the consistent sparse-positive case. In-region, full fine-tuning beats the
frozen decoder for both GeoFMs in the ten-seed grid: Prithvi frozen 0.653 vs.\ full-FT
0.771 ($\Delta=-0.117$, CI $[-0.176,-0.059]$), and TerraMind frozen 0.613 vs.\
full-FT 0.710 ($\Delta=-0.097$, CI $[-0.146,-0.048]$). Cross-region,
Kenya-involved transfers are also the hardest subset: on the 10 Kenya-involved
directions, U-Net averages 0.793 AUROC, while frozen Prithvi and frozen TerraMind
average 0.669 and 0.666.
Two measured properties distinguish Kenya. Its true-field-pixel coverage in the training
masks is about 0.1\%, the lowest of any region; the next lowest is India at 0.3\%.
The released field-size summaries show similar shares of fields under 0.3 ha in
India (65\%) and Kenya (70\%), yet only Kenya exhibits this collapse. Parcel size alone
is therefore insufficient to explain the result. Cambodia has the highest
positive-pixel coverage (23\%) and remains stable. Kenya has 155 training chips, whereas
Cambodia has 107 and remains stable, so training-set size alone is also insufficient.
Spatial fragmentation may contribute, but we did not quantify it independently.
Under this measured combination, frozen-feature decoders are unstable (seed standard
deviation up to $0.066$ AUROC), and the supervised U-Net is the most robust tested model on
threshold metrics.

\paragraph{Reconciling Kenya numbers across tables.} Kenya is the one region where
the three configurations defined in Section~\ref{sec:setup} disagree numerically.
\textbf{Best frozen-probe} (linear classifier on AnySat features, reported in
Appendix Table~\ref{tab:fm_breakdown} from a separate frozen-probe sensitivity,
not the ten-seed decoder grid) scores 0.853 AUROC;
\textbf{frozen-decoder} (Prithvi
features, 2.75M-parameter decoder, Table~\ref{tab:spine}) scores 0.653; the ten-seed threshold-metrics mean (Appendix Table~\ref{tab:xmetrics}) is 0.653 AUROC and
0.000 IoU.
Tile-disjoint sensitivity (Appendix Table~\ref{tab:tilesplit}) further shows that
AnySat's chip-grouped Kenya score does not survive a split-design change
($0.853 \to 0.580$), so this best-of-four Appendix cell is fragile in addition
to being from a different configuration. One possible explanation is that the generic
pretrained representation does not expose the features needed in this sparse regime. A
frozen backbone cannot adapt those features, whereas full fine-tuning and a from-scratch
U-Net can; they score 0.771 and 0.891 AUROC, respectively, compared with 0.653 for the
frozen decoder. This is a hypothesis, not a causal result. It also argues against
explaining the failure only by decoder size.
In Section~\ref{sec:cost} and the cross-region
narrative, we use the frozen-decoder Prithvi number (0.653) as the canonical Kenya
frozen value, since it is the apples-to-apples comparator against fine-tuning. Threshold metrics sharpen the boundary: at probability
threshold 0.5, the frozen decoder predicts all negatives and full fine-tuning is
near all-negative on Kenya (mean IoU 0.000 and 0.005; Appendix
Table~\ref{tab:xmetrics}); only the supervised U-Net recovers substantial positives
(mean IoU 0.086). The Kenya failure appears in both AUROC and threshold metrics,
so it is not an artifact of relying on AUROC alone.

We do not infer a hard sparsity threshold from one region. When annotated positives are
very sparse and the training set is small, the results instead motivate a direct comparison
between frozen decoders and a supervised CNN. Appendix Table~\ref{tab:regional_regime}
places Kenya in the context of the six regions.

A Cambodia subsampling control suggests that sample size alone does not explain the
failure. Reducing its full-fine-tuning set to 11, 27, or 54 chips, all far below Kenya's
155, leaves AUROC at
0.940--0.949, compared with 0.949 on all 107 chips, and its thresholded F1 does not
collapse (data in \texttt{ftw\_finetune\_fm\_prithvi\_camld\_*.json}). Nor is the
frozen-vs-fine-tuning difference monotone in positive-pixel prevalence across the other
regions. The evidence therefore supports describing Kenya as a joint sparse-positive and
small-training-set case, but it does not separate the two factors causally. A controlled
study that varies positive prevalence while holding chip count fixed remains future work.

\FloatBarrier
\section{Cost-Efficiency of Frozen Features}
\label{sec:cost}
Table~\ref{tab:cost} contrasts trainable parameters and per-region training cost. A
Prithvi frozen backbone with a decoder trains 2.75M parameters versus 307M for
Prithvi full fine-tuning, about $110\times$ fewer. Across both GeoFMs, frozen decoders are
TOST-equivalent to full fine-tuning in 9 of 12 region$\times$model cells
(Table~\ref{tab:paired_tost}). TerraMind India lies at the equivalence margin by point
estimate ($\Delta{=}{+}0.020$), but its CI extends beyond the margin. Kenya is the only region where full-FT
decisively wins for both GeoFMs. A frozen linear
probe trains about $10^3$ parameters in seconds on CPU. AUROC parity with the
31M-parameter U-Net is region-dependent: in the ten-seed in-region decoder comparison,
U-Net has higher point AUROC than frozen-Prithvi in four of six regions
(Cambodia, Vietnam, Kenya, Netherlands), while frozen-Prithvi is higher in India
and narrowly in France. The largest gap is Kenya, where U-Net leads frozen-Prithvi
by 0.237 AUROC. On IoU, the U-Net wins outright outside Cambodia
(Table~\ref{tab:iou_summary}). The frozen-probe cost advantage is robust;
matched-accuracy equality vs.\ the supervised baseline is region-specific. For
cross-region deployment, the frozen decoder also needs no target-region labels, but
the full transfer matrix shows that it should be compared directly with a supervised
U-Net; its transfer strength cannot be assumed.

\begin{figure}[!htb]
\centering
\includegraphics[width=0.62\linewidth]{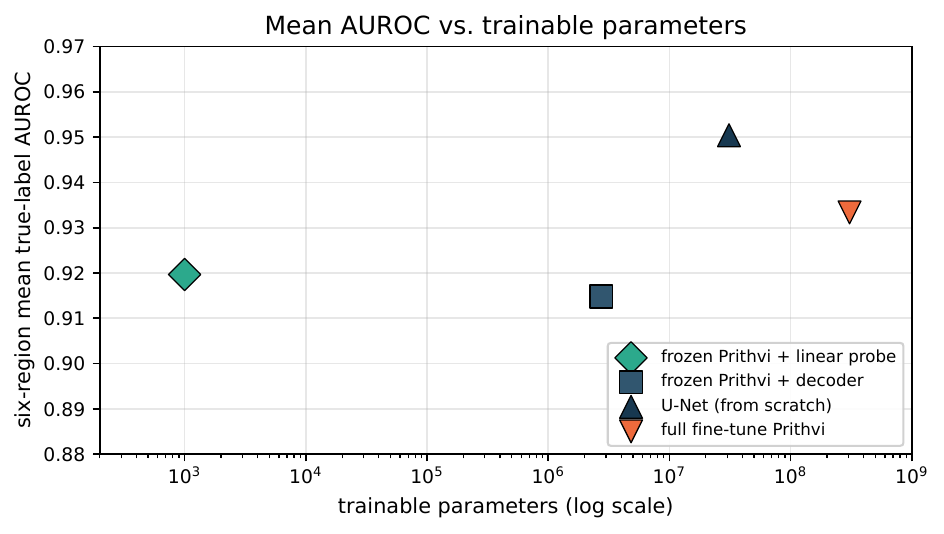}
\caption{\textbf{Cost vs.\ performance.} Six-region mean in-region AUROC against
trainable parameters (log scale). All GeoFM points use Prithvi consistently.
Frozen options (a $\sim$1K-parameter linear probe or a 2.75M-parameter decoder)
sit far left of the fully trained models. The frozen \emph{decoder} has
2.75M trainable parameters for Prithvi, compared with 307M for Prithvi full-FT. Across
both GeoFMs, frozen decoders are TOST-equivalent to full-FT in 9 of 12
region$\times$model cells (Table~\ref{tab:paired_tost}). The frozen \emph{probe} has lower point AUROC
than the 31M from-scratch U-Net in most regions; India is the main GeoFM
exception.}
\label{fig:cost}
\end{figure}

\begin{table}[!htb]
\centering\small
\caption{\textbf{Cost asymmetry.} Trainable parameters and per-region training cost.
The frozen-decoder pipeline is TOST-equivalent to full fine-tuning in 9 of 12
region$\times$model cells (Table~\ref{tab:paired_tost}). For Prithvi, the frozen
decoder trains a small fraction of the parameters used by full fine-tuning. GPU-minute entries are approximate development-run
observations; elapsed time was not retained in the result JSONs. AUROC
parity with the from-scratch U-Net is region-dependent; on IoU, the U-Net wins
outright outside Cambodia (Table~\ref{tab:iou_summary}).}
\label{tab:cost}
\tablebodygap
\begin{tabular}{lccc}
\toprule
Method & trainable params & per-region train & target labels \\
\midrule
per-pixel RF              & N/A (fitted trees) & seconds, CPU & yes \\
frozen FM + linear probe  & $\sim$1{,}000      & seconds, CPU & yes (source only, transfer) \\
frozen Prithvi + decoder  & 2.75M              & $\sim$12 GPU-min & yes \\
supervised U-Net          & 31M                & $\sim$15 GPU-min & yes \\
full fine-tune (Prithvi)  & 307M               & $\sim$20 GPU-min & yes \\
\bottomrule
\end{tabular}
\end{table}

\paragraph{Cost accounting: what is and is not counted.} Table~\ref{tab:cost} reports
trainable parameters and approximate per-region wall-clock observations from a
single A6000 GPU. Because elapsed time was not stored in the result JSONs, these
timings indicate scale but are not part of the reproducible quantitative claims.
The auditable cost statements are precisely scoped to two facts. First, the Prithvi
frozen decoder trains ${\sim}110\times$ fewer parameters than Prithvi full-FT
(2.75M vs.\ 307M). Second, for both GeoFMs, the frozen-decoder and full-FT
configurations run the identical backbone$+$decoder graph at inference,
so per-scene inference cost is equal; frozen does not reduce deployment-time backbone
compute. The released frozen-decoder pipeline does not cache features offline: the
frozen backbone still runs a forward pass every training epoch, so there is no
per-epoch compute saving from cached features and no cached-feature storage term to
report. A user who chooses to cache frozen features offline would pay a one-time
extraction and storage cost and could then reuse those features across multiple task
heads, but our released pipeline does not do this and therefore does not realize that
multi-head-reuse saving.

\FloatBarrier
\section{A Corrected-Evaluation Protocol}
\label{sec:protocol}
We distill the study into a ten-point checklist for field-level GeoFM benchmarking.
\begin{enumerate}
\item \textbf{Labels.} Evaluate against field-extent labels derived from field
polygons, not a land-cover proxy; treat proxy-only rankings as tentative.
\item Compare label sources on the same pixels.
\item \textbf{Baselines.} Include a strong task-specific supervised baseline
such as a U-Net alongside any per-pixel spectral model.
\item Report both a frozen-feature configuration and a fine-tuned variant; test
whether fine-tuning helps.
\item \textbf{Splits.} Group examples by chip.
\item Add a scene- or tile-disjoint split and a cross-region split.
\item \textbf{Preprocessing.} Use each model's own normalization and exclude
no-data pixels.
\item \textbf{Statistics.} Cluster uncertainty at the chip level, use multiple
seeds, and correct for multiple comparisons.
\item \textbf{Interpretation.} Run an edge-vs-interior control before claiming
boundary understanding.
\item Report field-size and label-sparsity regimes, trainable parameters, and
compute cost alongside accuracy. In this study, Kenya is the only region below
approximately 0.2\% positive pixels; this is a descriptive warning regime rather
than a general threshold.
\end{enumerate}

\paragraph{Worked example: re-reading a published ranking.} The Prithvi-EO-2.0 report
\citep{prithvi} reports a $+8.1$ mIoU gap for Prithvi-EO-2.0-600M over a
from-scratch U-Net on multi-temporal crop segmentation (50.7 vs.\ 42.6 mIoU in
its Table 9). Protocol step (1) flags direct transfer of that ranking to our
setting as task-dependent: crop-class segmentation is not polygon-derived
field-extent segmentation, so the result should be re-tested on field-polygon
labels before being treated as evidence about field membership. With
polygon-derived labels and a
strong baseline (steps 1 to 3), PANGAEA \citep{pangaea} reports on AI4SmallFarms
\citep{ai4smallfarms} (manually digitized parcels) a from-scratch U-Net at 46.3
mIoU, beating every evaluated GeoFM (Prithvi 26.9, DOFA 27.1, Scale-MAE 21.5).
Multi-temporal crop segmentation and single-date AI4SmallFarms parcel
delineation are different tasks/datasets, so this is not a direct apples-to-apples
inversion of the Prithvi result; rather, it illustrates why proxy-label wins
should be re-tested under steps (1)--(3).

\paragraph{Practitioner implications.} For Prithvi, a frozen backbone with a decoder
trains about $110\times$ fewer parameters than full fine-tuning; a linear probe
uses roughly $10^3$ trainable parameters (Table~\ref{tab:cost}). The released frozen-decoder pipeline does not
cache features: it still runs the backbone forward pass during training, and it uses
the same backbone$+$decoder graph as full-FT at inference. Users who explicitly cache
features offline can trade one-time extraction and storage for multi-head reuse, but
that saving is not realized in the reported runs.

For a new region with no local labels, the exploratory non-Kenya transfer subset
supports frozen decoders over full fine-tuning for both GeoFMs, but the full
cross-region matrix does not support a broad claim
that frozen GeoFMs beat a source-trained U-Net. On the full 30-transfer matrix the
U-Net has the highest AUROC and AP; on exploratory non-Kenya transfers no AUROC
difference is detected between TerraMind frozen and the U-Net, while Prithvi frozen is lower.

For this
single-date task, a frozen decoder is a low-cost configuration to test first, not
a general replacement for full fine-tuning or a substitute for a strong
supervised baseline; sparse-positive regimes require a direct comparison with a
supervised CNN.

\FloatBarrier
\section{Discussion, Limitations, Conclusion}
\label{sec:disc}
\paragraph{Scope.} Our claims are scoped to agricultural field-extent segmentation
from single-date Sentinel-2 with true field labels (per-pixel inside-vs-outside
field polygon; we do not evaluate boundary delineation or instance separation). They do not necessarily extend to
other tasks (e.g.\ crop-type, change detection), other sensors, or multi-temporal
regimes. GeoFM pretraining may help there.

\paragraph{Broader impact.} Automatic field-extent maps are used downstream in
agricultural planning, subsidy allocation, crop monitoring, and land-tenure
assessment. The same maps can also enable land-use surveillance, regulatory
enforcement, or asset valuation that affects smallholder farmers without their
consent. Our methodology choices (polygon-derived labels instead of land-cover proxies, a
strong supervised baseline, explicit reporting of boundary-zone failure) aim to
keep claims about ``what works'' aligned with the actual task and to make weakness
in sparse-label regimes (e.g.\ Kenya) explicit. Deployers should
not generalize results across agro-climatic regimes without re-evaluating on
local labels, and decisions with consequences for individuals should not be
automated from these maps alone.

\paragraph{Scope of the claim.} The claim is not that GeoFMs are useless. Current
evaluations overstate their advantage for in-distribution field mapping with
reasonable supervision; the practical value here is as frozen feature extractors,
which are low-cost and often match full fine-tuning in-region. Cross-region, the
value is narrower: frozen decoders beat full fine-tuning on the exploratory non-Kenya transfer
subset, but the U-Net has the highest AUROC and AP on the full directed matrix.

\paragraph{Limitations.} Our model coverage includes four GeoFMs and one supervised
architecture. PANGAEA evaluates more GeoFMs and reports a similar U-Net pattern, but
it does not replace broader testing within our setup. FTW annotations can also be
partial in sparse regions. A three-region sensitivity check does not identify that
partial annotation as the source of the frozen-FM result (Appendix Table~\ref{tab:partial}), but it cannot
rule out every labeling effect. Finally, the cross-region AUROC evidence is deliberately
narrow: the U-Net leads the full 30-transfer matrix, and the statistically supported
frozen-over-FT result is limited to the exploratory non-Kenya subset.

\paragraph{Training budgets.} The U-Net-vs-GeoFM comparisons are
architecture-appropriate but not compute-matched: the U-Net uses Adam for 150 epochs
from random initialization while the GeoFM frozen-decoder and full-FT runs use AdamW
for 80 epochs from pretrained weights. An 80-epoch U-Net control (Appendix
Table~\ref{tab:unet_epoch_sensitivity}, 3 seeds per region) preserves the qualitative
pattern of Table~\ref{tab:flaws} within 0.01 AUROC in five of six regions; India
shows a $0.03$ drop that widens the existing India frozen-probe lead but does not
create a new region-level exception. The frozen-decoder-vs-full-FT primary comparison
is separately budget-matched by construction (Table~\ref{tab:recipe}). Under
our approximate development-run timings, the U-Net is cheaper than full-FT per
region despite its longer schedule, although elapsed time was not retained in the
result JSONs. A per-epoch
convergence diagnostic (Appendix Figure~\ref{fig:convergence}) shows that the Vietnam
frozen-decoder curve changes by about $0.002$ AUROC from epoch 15 to 150, while the
Vietnam full-FT curve peaks near epoch 90 and then declines slightly. The Kenya full-FT
curve peaks at epoch 15 and ends lower at epoch 150, despite transient rebounds. No
configuration shows a sustained gain beyond the 80-epoch budget.

\paragraph{Boundary and cost controls.} The boundary-zone control
(Table~\ref{tab:edgezone}) is limited to
the frozen-probe configurations; extending it to the U-Net, frozen-decoder, and
full-FT configurations would require regenerating per-pixel predictions for those
trained models. We therefore report Table~\ref{tab:edgezone} strictly as a
frozen-probe scope check. Feature-extraction wall-time is not separately tabulated
because the frozen-decoder configuration used in the frozen-vs-fine-tuning spine does
not cache features to disk; extraction is folded into the training wall-time
(Table~\ref{tab:cost}). The table's GPU-minute values are approximate, unlogged
observations; the auditable cost comparison is the trainable-parameter count.

\paragraph{Connection to the broader foundation-model literature.} For geospatial models, the cross-region result is partly consistent with what
\citet{kumar2022finetuning} showed for vision: fine-tuning can distort pretrained
features and underperform frozen probing under distribution shift. We see that
pattern on the exploratory non-Kenya transfer subset, where frozen decoders beat full
fine-tuning for both GeoFMs. The full 30-transfer matrix is more cautious because
Kenya-involved transfers favor task-trained supervised models and the U-Net has the
highest AUROC and AP overall. This supports a narrower recommendation: test a frozen decoder
before full fine-tuning, but do not treat it as a transfer rule. The released
evaluation setup can test where this pattern persists.

\paragraph{Conclusion.} Two evaluation choices materially change this field-extent
benchmark. WorldCover cropland makes the target more spectrally separable than
polygon-derived field extent, and a task-trained U-Net is much stronger than a per-pixel
random forest. GeoFM rankings should therefore be interpreted only after the label and
baseline choices are made explicit.

In the controlled ten-seed comparison, frozen decoders are TOST-equivalent to full
fine-tuning in 9 of 12 region$\times$model cells. For Prithvi, the frozen-decoder
configuration trains about $110\times$ fewer parameters. Kenya is the shared failure case for both GeoFMs. Cross-region, the U-Net leads
the full 30-transfer matrix on AUROC and AP. On AUROC, frozen decoders outperform full fine-tuning only in the
exploratory subset that excludes Kenya, and no difference is detected between frozen
TerraMind and the U-Net in AUROC there. The practical recommendation is to test a frozen decoder
before paying for full fine-tuning, while retaining a strong supervised baseline and
checking sparse-positive regions directly. The released code, metrics, and protocol support
the same audit in other GeoFM evaluations. These conclusions apply to single-date,
per-pixel field-extent segmentation, not parcel-boundary or instance delineation;
multi-temporal regimes, where Prithvi-EO-2.0 was designed to operate, are out of scope.

\section*{Acknowledgments}
The author thanks Knox College for providing the computational resources used in this
work. This research received no external funding.

\bibliographystyle{tmlr}
\bibliography{references}

\appendix
\FloatBarrier
\raggedbottom
\section{Reproducibility and Controls}
\label{app:repro}

\begin{figure}[H]
\centering
\includegraphics[width=0.9\linewidth]{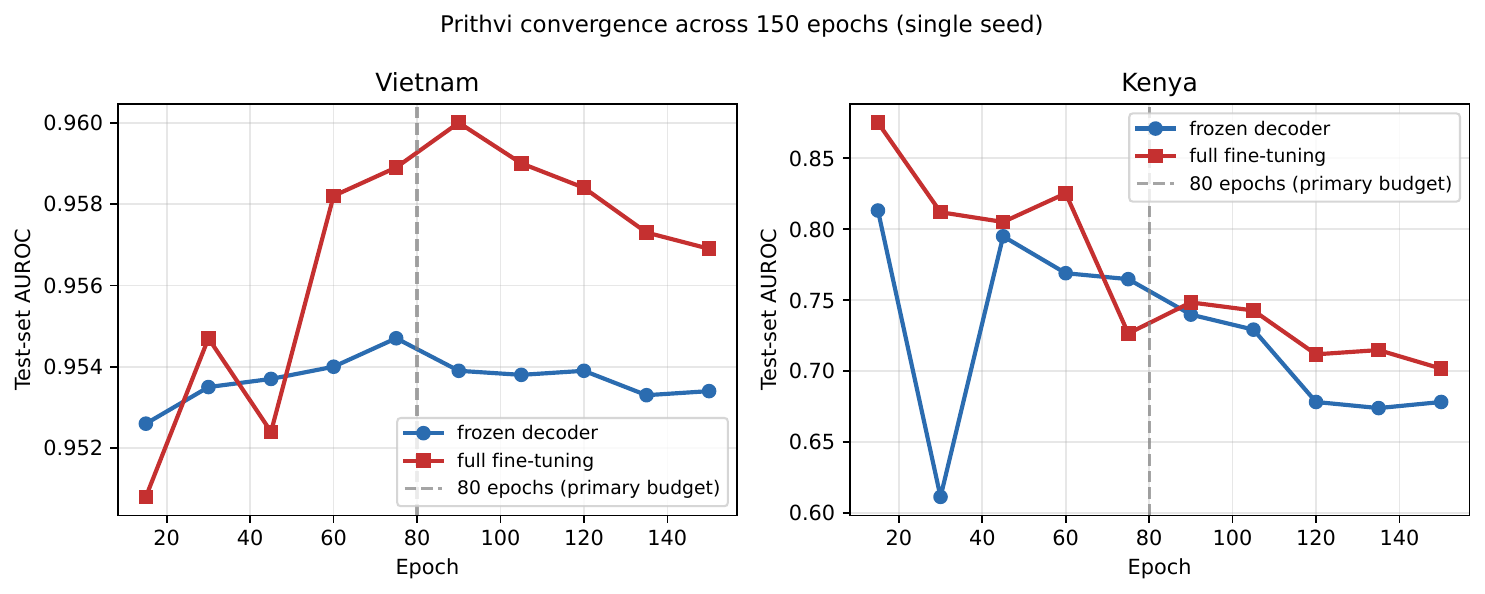}
\caption{\textbf{Prithvi test-set AUROC across 150 epochs (single-seed diagnostic).}
Post-hoc test-set AUROC for Prithvi frozen-decoder and full
fine-tuning on Vietnam (dense-positive, $18\%$ positives) and Kenya (sparse-positive case,
$0.1\%$ positives). Vertical dashed line at epoch $80$ marks the primary-comparison
budget. \emph{Vietnam:} the frozen-decoder curve changes by about $0.002$ AUROC from
epoch 15 through 150; the full-FT curve peaks near epoch 90 (${\approx}0.960$)
and slightly declines to $0.957$ at epoch 150. \emph{Kenya:} the full-FT curve peaks
at epoch 15 ($0.875$) and declines to $0.702$ by epoch 150, with small transient rebounds around epochs 60 and 90; the
frozen-decoder is unstable and never recovers a stable value in this regime, ending
at $0.678$. No configuration shows a sustained gain beyond the 80-epoch budget, and
both Kenya curves end lower at epoch 150 despite transient rebounds. These post-hoc
trajectories were not used to choose checkpoints or hyperparameters; model selection
remains last-epoch with no early stopping.}
\label{fig:convergence}
\end{figure}

This appendix provides the detailed results and controls referenced in the main text.
Tables~\ref{tab:fm_breakdown}--\ref{tab:recipe} cover the frozen-probe breakdown,
paired TOST analysis, margin sensitivity, and training recipes.
Tables~\ref{tab:xfer-full}--\ref{tab:xfer_metrics} report the full transfer matrix
and its AP/IoU companion. The remaining tables and Figure~\ref{fig:convergence}
report budget, regional, spatial, partial-label, split, and threshold-metric controls.
Unless a caption states otherwise, values are generated by the released scripts from
the released result files; rounded mask-prevalence descriptors are recorded in
\verb|scripts/integrate_revision_controls.py| because source masks are not redistributed.

\begin{table}[H]\centering\small
\caption{\textbf{Per-region breakdown of frozen-probe AUROC across four FMs}
(true-label, linear probe, deterministic given the split). The Table~\ref{tab:flaws}
main column uses Prithvi-EO-2.0 consistently; this table reports each FM separately
for transparency. \textbf{Bold} marks the per-region winner. AnySat wins Kenya
(0.853) and Netherlands (0.961), but tile-disjoint sensitivity
(Table~\ref{tab:tilesplit}) shows the AnySat scores do not survive a split-design
change ($0.853\to0.580$ on Kenya; $0.943\to0.605$ on France); we therefore use the
Prithvi-only column as the main comparison. Bootstrap 95\% CIs from $B{=}1000$ chip-level
resamples.}
\label{tab:fm_breakdown}
\tablebodygap
\begin{tabular}{lcccc}
\toprule
Region & Prithvi & TerraMind & Clay & AnySat \\
\midrule
India        & \textbf{0.983}\,{\scriptsize[.979,.985]} & 0.967\,{\scriptsize[.963,.971]} & 0.577\,{\scriptsize[.554,.601]} & 0.939\,{\scriptsize[.915,.959]} \\
Vietnam      & \textbf{0.929}\,{\scriptsize[.925,.933]} & 0.916\,{\scriptsize[.911,.921]} & 0.503\,{\scriptsize[.495,.511]} & 0.897\,{\scriptsize[.880,.915]} \\
Kenya        & 0.766\,{\scriptsize[.718,.813]} & 0.745\,{\scriptsize[.699,.789]} & 0.556\,{\scriptsize[.505,.599]} & \textbf{0.853}\,{\scriptsize[.825,.881]} \\
France       & \textbf{0.961}\,{\scriptsize[.958,.964]} & 0.951\,{\scriptsize[.947,.954]} & 0.501\,{\scriptsize[.494,.509]} & 0.943\,{\scriptsize[.927,.957]} \\
Netherlands  & 0.955\,{\scriptsize[.951,.958]} & 0.943\,{\scriptsize[.938,.948]} & 0.498\,{\scriptsize[.488,.507]} & \textbf{0.961}\,{\scriptsize[.956,.967]} \\
\bottomrule
\end{tabular}
\\[2pt]
{\scriptsize Cambodia per-chip bootstrap CIs were not recomputed; available
per-model point estimates are Prithvi 0.924, TerraMind 0.918, and Clay 0.491.
No Cambodia AnySat result is present in the artifact.}
\end{table}

\begin{table}[H]\centering\tiny
\setlength{\tabcolsep}{2pt}\renewcommand{\arraystretch}{0.93}
\caption{\textbf{Seed-paired TOST equivalence for frozen-decoder vs.\ full
 fine-tune (12 cells: two \mbox{GeoFMs} $\times$ six regions).} Per-seed AUROC values
(seeds 0--9), paired delta $\Delta_s = \mathrm{frozen}_s - \mathrm{FT}_s$ averaged
over seeds, 95\% paired $t$-interval with $\mathrm{df}=9$, and TOST verdict at declared
equivalence margin $\varepsilon{=}0.02$. The paired analysis uses matched random
seeds for each frozen/FT pair in the single-recipe 10-seed grid. Source files
and aggregate JSON are generated by \texttt{scripts/integrate\_headline\_results.py}. Total
TOST-equivalent: 9 of 12.}
\label{tab:paired_tost}
\tablebodygap
\begin{tabular}{llp{0.23\linewidth}p{0.23\linewidth}cc}
\toprule
Model & Region & frozen seeds (s0--s9) & FT seeds (s0--s9) & paired $\Delta \pm \sigma$ & 95\% CI \\
\midrule
Prithvi & India & .984, .981, .983, .985, .983, .979, .980, .979, .985, .978 & .982, .988, .985, .986, .986, .986, .983, .982, .988, .986 & $-0.004 \pm 0.003$ & $[-0.006,-0.002]$ \checkmark \\
Prithvi & Cambodia & .948, .948, .946, .947, .946, .946, .948, .948, .948, .947 & .949, .949, .948, .950, .949, .949, .949, .949, .949, .948 & $-0.002 \pm 0.001$ & $[-0.003,-0.001]$ \checkmark \\
Prithvi & Vietnam & .954, .955, .954, .955, .955, .955, .955, .954, .955, .955 & .958, .957, .958, .959, .959, .959, .958, .958, .958, .960 & $-0.004 \pm 0.001$ & $[-0.004,-0.003]$ \checkmark \\
Prithvi & Kenya & .667, .658, .615, .599, .607, .676, .531, .723, .745, .713 & .781, .798, .699, .811, .814, .728, .761, .701, .791, .824 & $-0.117 \pm 0.082$ & $[-0.176,-0.059]$ \textbf{$\times$} \\
Prithvi & France & .981, .982, .981, .982, .982, .982, .982, .981, .981, .981 & .980, .978, .981, .979, .979, .977, .977, .979, .979, .979 & $+0.003 \pm 0.002$ & $[+0.002,+0.004]$ \checkmark \\
Prithvi & Netherlands & .971, .973, .971, .971, .970, .974, .971, .968, .968, .971 & .964, .957, .953, .954, .963, .961, .946, .964, .951, .965 & $+0.013 \pm 0.007$ & $[+0.008,+0.018]$ \checkmark \\
\midrule
TerraMind & India & .972, .934, .976, .977, .973, .975, .978, .964, .970, .965 & .970, .955, .950, .946, .961, .956, .946, .930, .912, .955 & $+0.020 \pm 0.021$ & $[+0.005,+0.036]$ \textbf{$\times$} \\
TerraMind & Cambodia & .943, .942, .943, .942, .941, .943, .943, .942, .942, .942 & .949, .948, .947, .949, .949, .948, .949, .949, .949, .947 & $-0.006 \pm 0.001$ & $[-0.007,-0.005]$ \checkmark \\
TerraMind & Vietnam & .951, .951, .952, .951, .951, .951, .952, .951, .951, .951 & .955, .956, .956, .956, .956, .956, .956, .956, .955, .956 & $-0.004 \pm 0.000$ & $[-0.005,-0.004]$ \checkmark \\
TerraMind & Kenya & .610, .632, .638, .596, .537, .576, .707, .643, .569, .625 & .820, .724, .788, .611, .573, .696, .696, .773, .721, .703 & $-0.097 \pm 0.069$ & $[-0.146,-0.048]$ \textbf{$\times$} \\
TerraMind & France & .978, .979, .979, .979, .979, .979, .979, .979, .979, .978 & .979, .978, .978, .978, .977, .979, .975, .978, .978, .978 & $+0.001 \pm 0.001$ & $[+0.000,+0.002]$ \checkmark \\
TerraMind & Netherlands & .960, .958, .958, .956, .964, .965, .964, .960, .964, .959 & .942, .951, .961, .953, .943, .964, .962, .940, .954, .952 & $+0.009 \pm 0.009$ & $[+0.003,+0.015]$ \checkmark \\
\bottomrule
\end{tabular}
\\[2pt]
{\footnotesize \checkmark\,=\,TOST equivalent at $\varepsilon{=}0.02$ AUROC (95\%
paired CI fully within $\pm 0.02$; equivalent to standard TOST at $\alpha{=}0.025$
per side, conservative form). $\times$\,=\,not formally equivalent. Kenya is a
full-FT win for both GeoFMs; TerraMind India is frozen-favored but its CI extends
past $+0.02$.}
\end{table}

\begin{table}[H]
\centering\small
\caption{\textbf{Equivalence sensitivity to the TOST margin.}
Table counts use the same 95\% seed-paired CI-containment criterion as the primary verdicts
($\mathrm{df}=9$). At five of the six margins, Holm correction across the 12 TOST
$p$-values retains the same number of cells. At $\varepsilon{=}0.015$, it retains 7
rather than 8 because TerraMind--Netherlands has
$p_{\mathrm{Holm}}{=}0.112$. The failing set is
invariant across $\varepsilon \in [0.020, 0.030]$. Cells added at stricter
margins in $[0.010, 0.020)$ are frozen-favored (F), so tightening the margin in
this range only reclassifies cells where the frozen decoder already beats fine-tuning.
At $\varepsilon{=}0.005$ two additional cells become non-equivalent
in the FT-favored direction (T), but their $\Delta$ magnitudes ($0.004$ and $0.006$)
are below the declared practical tolerance.}
\label{tab:eps_sweep}
\tablebodygap
\footnotesize
\setlength{\tabcolsep}{3pt}
\begin{tabular}{@{}rcp{0.29\linewidth}p{0.39\linewidth}@{}}
\toprule
$\varepsilon$ & Equivalent & Newly failing (direction, $\Delta$) & Failing cells at this $\varepsilon$ \\
\midrule
$0.030$ & $9/12$ & none & \{Pr--Kenya (T), TM--India (F), TM--Kenya (T)\} \\
$0.025$ & $9/12$ & none & \{same\} \\
$0.020$ & $9/12$ & none & \{same\} \\
$0.015$ & $8/12$ & Pr--Netherlands (F, $+0.013$) & \{above\} $\cup$ Pr--Netherlands \\
$0.010$ & $7/12$ & TM--Netherlands (F, $+0.009$) & \{above\} $\cup$ TM--Netherlands \\
$0.005$ & $5/12$ & Pr--India (T, $-0.004$); TM--Cambodia (T, $-0.006$) & \{above\} $\cup$ Pr--India, TM--Cambodia \\
\bottomrule
\end{tabular}
\end{table}

All experiments use the same chip-grouped split (seed 20260514; no chip overlap).
Primary results recompute from the released run files, and direct raster checks verify
label-to-image alignment in all six regions. The released controls cover spatial
pooling, frozen-probe edge behavior, partial labels, and tile-disjoint splits. The
single-recipe ten-seed grid includes all 30 directed transfers for U-Net, Prithvi,
and TerraMind. Historical pilot files are retained for traceability but are not used
for the paper's claims. Code, checksums, reproduction commands, and Croissant metadata
are included in the supplement and at
\url{https://github.com/thesantoshpant/auditing-geofm-evaluation} and archived at
\url{https://doi.org/10.5281/zenodo.21993508} (v1.0.0).

\begin{table}[!htb]\centering\small
\caption{\textbf{Training recipes and controlled factors.} Frozen-decoder and
full-FT GeoFM runs share the same decoder, loss, augmentation, schedule, and matched
seeds; they differ only in whether the backbone is updated. The U-Net uses its
architecture-appropriate optimizer and longer from-scratch schedule. Evaluation
pixels, labels, and the chip-grouped split are identical for every model and seed.
The U-Net and GeoFM training budgets differ by design (150 ep from-scratch vs.\ 80 ep
from pretrained), matching common practice for each architecture. An 80-epoch U-Net
sensitivity (Appendix Table~\ref{tab:unet_epoch_sensitivity}) and a per-epoch
convergence diagnostic (Appendix Figure~\ref{fig:convergence}) are provided for audit
purposes.}
\label{tab:recipe}
\tablebodygap
\begin{tabular}{l p{0.72\linewidth}}
\toprule
Component & Setting \\
\midrule
Loss            & BCE (capped positive weight, $\le 50$) $+$ Dice \\
Optimizer       & U-Net: Adam, lr $10^{-3}$;\ \ GeoFM: AdamW (backbone $10^{-4}$, decoder $10^{-3}$) \\
LR schedule     & linear warmup (5 epochs, start factor 0.1) $+$ cosine decay \\
Augmentation    & random horizontal/vertical flips \\
Input           & U-Net: native chip, 12 S2 bands (reflect-padded); GeoFM: 224$\times$224 \\
Decoder         & GeoFM: shared 4-stage bilinear-upsample conv decoder (identical for frozen \& fine-tuned); U-Net: 4-level (31M) \\
Epochs          & U-Net 150 (from scratch); GeoFM full-FT / frozen-decoder 80 \\
Selection       & last epoch (no early stopping); mean$\pm$std over 10 matched training seeds in the ten-seed grid \\
\bottomrule
\end{tabular}
\end{table}

\begin{table}[H]\centering\scriptsize
\setlength{\tabcolsep}{3pt}\renewcommand{\arraystretch}{0.85}
\caption{\textbf{Full cross-region transfer matrix} (true-label AUROC, mean
over 10 seeds; train country A, test country B's canonical test pixels). The table
covers all 30 directed transfers among the six regions. U-Net has the highest mean AUROC
on the full matrix (0.826). On the exploratory non-Kenya 20 subset, TerraMind frozen-decoder (0.839) is
close to U-Net (0.842) and beats TerraMind full-FT (0.813), while Prithvi frozen-decoder
(0.816) beats Prithvi full-FT (0.810) but remains below U-Net. Seed-paired CIs for
subset-level deltas are generated in \texttt{data/results/ftw\_cross\_region\_transfer.json}.
In the compact headers, ``frz'' denotes frozen-decoder.}
\label{tab:xfer-full}
\tablebodygap
\begin{tabular}{lccccc}
\toprule
A$\rightarrow$B & U-Net & Prithvi-FT & Prithvi-frz & TerraMind-FT & TerraMind-frz \\
\midrule
India$\to$Cambodia & 0.719 & 0.646 & 0.629 & 0.627 & 0.697 \\
India$\to$Vietnam & 0.698 & 0.664 & 0.653 & 0.643 & 0.699 \\
India$\to$Kenya & 0.696 & 0.629 & 0.579 & 0.546 & 0.616 \\
India$\to$France & 0.824 & 0.750 & 0.644 & 0.611 & 0.660 \\
India$\to$Netherlands & 0.830 & 0.746 & 0.675 & 0.614 & 0.646 \\
Cambodia$\to$India & 0.832 & 0.861 & 0.890 & 0.870 & 0.885 \\
Cambodia$\to$Vietnam & 0.874 & 0.926 & 0.922 & 0.940 & 0.917 \\
Cambodia$\to$Kenya & 0.881 & 0.836 & 0.865 & 0.862 & 0.853 \\
Cambodia$\to$France & 0.865 & 0.887 & 0.910 & 0.920 & 0.884 \\
Cambodia$\to$Netherlands & 0.895 & 0.883 & 0.884 & 0.901 & 0.879 \\
Vietnam$\to$India & 0.873 & 0.868 & 0.870 & 0.875 & 0.886 \\
Vietnam$\to$Cambodia & 0.939 & 0.938 & 0.938 & 0.945 & 0.930 \\
Vietnam$\to$Kenya & 0.897 & 0.862 & 0.884 & 0.890 & 0.885 \\
Vietnam$\to$France & 0.925 & 0.907 & 0.919 & 0.930 & 0.931 \\
Vietnam$\to$Netherlands & 0.884 & 0.843 & 0.898 & 0.898 & 0.898 \\
Kenya$\to$India & 0.790 & 0.678 & 0.517 & 0.642 & 0.596 \\
Kenya$\to$Cambodia & 0.884 & 0.723 & 0.622 & 0.734 & 0.592 \\
Kenya$\to$Vietnam & 0.746 & 0.642 & 0.588 & 0.690 & 0.543 \\
Kenya$\to$France & 0.800 & 0.767 & 0.629 & 0.692 & 0.550 \\
Kenya$\to$Netherlands & 0.851 & 0.770 & 0.640 & 0.738 & 0.596 \\
France$\to$India & 0.887 & 0.877 & 0.903 & 0.891 & 0.904 \\
France$\to$Cambodia & 0.840 & 0.740 & 0.748 & 0.781 & 0.750 \\
France$\to$Vietnam & 0.835 & 0.688 & 0.742 & 0.794 & 0.828 \\
France$\to$Kenya & 0.761 & 0.672 & 0.722 & 0.714 & 0.735 \\
France$\to$Netherlands & 0.937 & 0.905 & 0.931 & 0.883 & 0.926 \\
Netherlands$\to$India & 0.864 & 0.856 & 0.898 & 0.871 & 0.881 \\
Netherlands$\to$Cambodia & 0.691 & 0.700 & 0.691 & 0.716 & 0.777 \\
Netherlands$\to$Vietnam & 0.692 & 0.619 & 0.627 & 0.648 & 0.834 \\
Netherlands$\to$Kenya & 0.625 & 0.618 & 0.642 & 0.663 & 0.692 \\
Netherlands$\to$France & 0.940 & 0.901 & 0.954 & 0.904 & 0.960 \\
\midrule
mean all 30 & 0.826 & 0.780 & 0.767 & 0.781 & 0.781 \\
mean expl.\ non-Kenya 20 & 0.842 & 0.810 & 0.816 & 0.813 & 0.839 \\
\bottomrule
\end{tabular}
\end{table}

\begin{table}[H]\centering\small
\caption{\textbf{Cross-region AP and IoU alongside AUROC (mean over 10 seeds per pair,
then mean over pairs).} Threshold metrics (IoU at $0.5$) and AP are reported in addition
to AUROC because low positive rates (Kenya $\sim0.1\%$, India $\sim0.3\%$) can make
AUROC alone optimistic. The principal AUROC conclusion is preserved under AP: U-Net
leads the full matrix, and U-Net and TerraMind frozen-decoder differ by only 0.001 AP on the
exploratory non-Kenya subset. The IoU@0.5 column is a
thresholded segmentation-quality metric; under this metric the ordering is
metric-dependent (see Section~\ref{sec:spine}). Aggregated from existing per-run JSONs;
no new experiments. See \texttt{data/results/ftw\_xfer\_metrics.json}.}
\label{tab:xfer_metrics}
\tablebodygap
\setlength{\tabcolsep}{4pt}
\begin{tabular}{@{}lcccccc@{}}
\toprule
& \multicolumn{3}{c}{all 30 directed transfers} & \multicolumn{3}{c}{exploratory non-Kenya 20} \\
\cmidrule(lr){2-4}\cmidrule(lr){5-7}
Config & AUROC & AP & IoU@0.5 & AUROC & AP & IoU@0.5 \\
\midrule
U-Net (scratch)   & \textbf{0.826} & \textbf{0.733} & 0.335 & \textbf{0.842} & \textbf{0.805} & 0.423 \\
TerraMind frozen-dec. & 0.781 & 0.696 & \textbf{0.353} & 0.839 & 0.804 & \textbf{0.469} \\
TerraMind full-FT & 0.781 & 0.693 & 0.318 & 0.813 & 0.778 & 0.416 \\
Prithvi full-FT   & 0.780 & 0.689 & 0.308 & 0.810 & 0.771 & 0.402 \\
Prithvi frozen-dec. & 0.767 & 0.682 & 0.308 & 0.816 & 0.782 & 0.404 \\
\bottomrule
\end{tabular}
\end{table}

\begin{table}[H]\centering\small
\caption{\textbf{U-Net epoch sensitivity.}
The 80-epoch control reports mean$\pm$standard deviation over seeds 0--2; the
150-epoch column reports the primary ten-seed mean. The comparison is a budget
sensitivity check, not a seed-paired test; architecture, optimizer, data, and schedule
family are unchanged. The displayed $\Delta$ is computed from unrounded means.
At 80 epochs the U-Net remains within $0.01$ AUROC of the 150-epoch mean in
five of six regions; India shows a $0.03$ drop but remains the same region-level
exception by point estimate. Training-time cost
per region is roughly proportional to epochs (Table~\ref{tab:cost}).}
\label{tab:unet_epoch_sensitivity}
\tablebodygap
\setlength{\tabcolsep}{3pt}
\begin{tabular}{@{}lcccc@{}}
\toprule
Region & U-Net@150ep & U-Net@80ep & $\Delta$ AUROC & Prithvi frozen-probe \\
\midrule
India        & 0.941 & $0.911 \pm 0.023$ & $-0.030$ & 0.983 \\
Cambodia     & 0.948 & $0.947 \pm 0.001$ & $-0.001$ & 0.924 \\
Vietnam      & 0.964 & $0.955 \pm 0.001$ & $-0.009$ & 0.929 \\
Kenya        & 0.891 & $0.889 \pm 0.004$ & $-0.002$ & 0.766 \\
France       & 0.977 & $0.976 \pm 0.001$ & $-0.001$ & 0.961 \\
Netherlands  & 0.981 & $0.982 \pm 0.003$ & $+0.000$ & 0.955 \\
\bottomrule
\end{tabular}
\end{table}

\begin{table}[H]\centering\small
\caption{\textbf{Per-region operational regime (descriptive, existing artifacts only).}
Positive-pixel prevalence and training-set size vary by orders of magnitude across the
six regions. Combined with the frozen-decoder-vs-full-fine-tuning $\Delta$AUROC (Prithvi
and TerraMind) and threshold-metric behavior (U-Net vs.\ frozen-decoder IoU@0.5), this
shows where frozen features remain competitive and where they fail in these six regions. Kenya is the
only region below $\sim0.2\%$ positive pixels, so we do not report a formal sparsity
threshold; the pattern is a description, not an estimate. Sample-size alone is not the
mechanism: subsampling Cambodia (dense-positive) full-fine-tuning to $11$--$54$ training
chips leaves AUROC essentially flat ($0.940$--$0.949$; \texttt{ftw\_finetune\_fm\_prithvi\_camld\_*}).
A controlled frozen-decoder low-data arm at Kenya-scale chip count is future work.}
\label{tab:regional_regime}
\tablebodygap
\footnotesize
\setlength{\tabcolsep}{3pt}
\begin{tabular}{@{}lrrrrccl@{}}
\toprule
Region & pos.\ \% & train chips & $\Delta$AUROC Pr & $\Delta$AUROC TM & U-Net IoU & Frozen IoU & Regime \\
\midrule
Kenya       & $\sim$0.1 & 155 & $-0.117$ & $-0.097$ & 0.086 & 0.000 & sparse-positive collapse \\
India       & $\sim$0.3 & 525 & $-0.004$ & $+0.020$ & 0.455 & 0.386 & smallholder; model-dependent \\
Netherlands & $\sim$3.8 & 522 & $+0.013$ & $+0.009$ & 0.809 & 0.699 & frozen-favored AUROC \\
France      & $\sim$11  & 532 & $+0.003$ & $+0.001$ & 0.891 & 0.850 & industrial; near-parity \\
Vietnam     & $\sim$18  & 213 & $-0.004$ & $-0.004$ & 0.854 & 0.839 & tie \\
Cambodia    & $\sim$23  & 107 & $-0.002$ & $-0.006$ & 0.826 & 0.850 & tie \\
\bottomrule
\end{tabular}
\end{table}

\begin{table}[H]\centering\small
\caption{\textbf{Spatial-pooling control (true-label AUROC).} A $5\times5$ windowed
random forest (300-d) gains only about 0.01 over the per-pixel RF on average (Kenya drops by the same amount) and stays far
below the best foundation model. This supports the interpretation that the GeoFM
and U-Net advantage uses learned spatial context, which naive local pooling lacks.
A per-pixel U-Net ablation
($1\times1$ convolutions, no spatial context) likewise falls to $\approx$RF
(India 0.623, Vietnam 0.678).}
\label{tab:windowed}
\tablebodygap
\begin{tabular}{lccc}
\toprule
Region & per-pixel RF & windowed RF (5$\times$5) & best FM \\
\midrule
India       & 0.574 & 0.582 & 0.983 \\
Vietnam     & 0.643 & 0.657 & 0.929 \\
Kenya       & 0.651 & 0.641 & 0.853 \\
France      & 0.695 & 0.703 & 0.961 \\
Netherlands & 0.815 & 0.827 & 0.961 \\
\bottomrule
\end{tabular}
\end{table}

\begin{table}[H]\centering\small
\caption{\textbf{FTW partial-labeling sensitivity (true-label AUROC).} Restricting
negatives to high-confidence non-cropland (dropping ambiguous, possibly-unlabeled
field pixels) leaves the frozen-GeoFM nearly unchanged in the three tested regions,
while the spectral RF rises. This control does not indicate that FTW partial
annotation drives the frozen-GeoFM result. France's RF gain
($0.695 \to 0.947$) reflects spectrally crop-like negatives (pasture, vineyards)
that confuse a per-pixel RF; removing them leaves an easy task. The GeoFM was
near-ceiling, so it barely moves.}
\label{tab:partial}
\tablebodygap
\begin{tabular}{lcccc}
\toprule
Region & RF (all neg) & RF (high-conf) & Prithvi (all neg) & Prithvi (high-conf) \\
\midrule
India  & 0.574 & 0.665 & 0.983 & 0.987 \\
France & 0.695 & 0.947 & 0.961 & 0.968 \\
Kenya  & 0.651 & 0.698 & 0.766 & 0.774 \\
\bottomrule
\end{tabular}
\end{table}

\begin{table}[H]\centering\small
\caption{\textbf{Tile-disjoint sensitivity for the Table~\ref{tab:flaws}
frozen-probe column (true-label AUROC).} Each pair: chip-grouped (used in the main
text) vs.\ MGRS-tile-disjoint AUROC for the same frozen probe. Prithvi-EO-2.0 is
stable across both splits (drops $\le 0.007$). TerraMind is also stable except in
Kenya, where it drops from 0.745 to 0.698. AnySat
drops on Kenya ($0.853 \to 0.580$), Vietnam ($0.897 \to 0.645$), and France
($0.943 \to 0.605$): its chip-grouped advantage may not reflect transferable feature
quality. In Kenya, Prithvi ($0.761$) is more robust under tile-disjoint than AnySat
($0.580$). Cambodia is omitted (variant not run). Chip-grouped numbers preserve
the per-chip statistical assumptions of the controlled comparison; this is the
sensitivity check.}
\label{tab:tilesplit}
\tablebodygap
\setlength{\tabcolsep}{7pt}
\begin{tabular}{lcccccc}
\toprule
        & \multicolumn{2}{c}{Prithvi-EO-2.0} & \multicolumn{2}{c}{TerraMind} & \multicolumn{2}{c}{AnySat} \\
\cmidrule(lr){2-3}\cmidrule(lr){4-5}\cmidrule(lr){6-7}
Region  & chip & tile & chip & tile & chip & tile \\
\midrule
India       & 0.983 & 0.984 & 0.967 & 0.969 & 0.939 & 0.945 \\
Vietnam     & 0.929 & 0.922 & 0.916 & 0.915 & 0.897 & 0.645 \\
Kenya       & 0.766 & 0.761 & 0.745 & 0.698 & 0.853 & 0.580 \\
France      & 0.961 & 0.959 & 0.951 & 0.947 & 0.943 & 0.605 \\
Netherlands & 0.955 & 0.955 & 0.943 & 0.944 & 0.961 & 0.893 \\
\bottomrule
\end{tabular}
\end{table}

\begin{table}[H]\centering\footnotesize
\setlength{\tabcolsep}{5pt}\renewcommand{\arraystretch}{0.95}
\caption{\textbf{Threshold-based metrics complement AUROC} (mean over 10 seeds;
F1/IoU at $p{=}0.5$; AP threshold-free). Dense-positive Cambodia and Vietnam show
the three configurations close on IoU. France and Netherlands show the U-Net
ahead on IoU, so the threshold metrics favor the supervised baseline more
than AUROC alone would suggest. Kenya ($\sim$0.1\% positives): frozen decoder
predicts all negatives (F1$=$IoU$=$0), full fine-tune is near all-negative, and U-Net retains
AP$=$0.593, F1$=$0.159.
``Frozen'' = Prithvi backbone frozen, decoder trained.}
\label{tab:xmetrics}
\tablebodygap
\begin{tabular}{llcccc}
\toprule
Region & Model & AUROC & AP & F1 & IoU \\
\midrule
India       & U-Net   & 0.941 & 0.856 & 0.619 & 0.455 \\
India       & Frozen  & 0.982 & 0.970 & 0.557 & 0.386 \\
India       & Full-FT & 0.985 & 0.974 & 0.604 & 0.433 \\
\midrule
Cambodia    & U-Net   & 0.948 & 0.923 & 0.904 & 0.826 \\
Cambodia    & Frozen  & 0.947 & 0.912 & 0.919 & 0.850 \\
Cambodia    & Full-FT & 0.949 & 0.914 & 0.923 & 0.856 \\
\midrule
Vietnam     & U-Net   & 0.964 & 0.948 & 0.921 & 0.854 \\
Vietnam     & Frozen  & 0.955 & 0.927 & 0.912 & 0.839 \\
Vietnam     & Full-FT & 0.958 & 0.933 & 0.917 & 0.847 \\
\midrule
Kenya       & U-Net   & 0.891 & 0.593 & 0.159 & 0.086 \\
Kenya       & Frozen  & 0.653 & 0.304 & 0.000 & 0.000 \\
Kenya       & Full-FT & 0.771 & 0.403 & 0.010 & 0.005 \\
\midrule
France      & U-Net   & 0.977 & 0.963 & 0.942 & 0.891 \\
France      & Frozen  & 0.982 & 0.972 & 0.919 & 0.850 \\
France      & Full-FT & 0.979 & 0.972 & 0.923 & 0.856 \\
\midrule
Netherlands & U-Net   & 0.981 & 0.977 & 0.895 & 0.809 \\
Netherlands & Frozen  & 0.971 & 0.967 & 0.823 & 0.699 \\
Netherlands & Full-FT & 0.958 & 0.962 & 0.829 & 0.708 \\
\bottomrule
\end{tabular}
\end{table}

\end{document}